\documentclass[runningheads]{llncs}
\usepackage[T1]{fontenc}
\usepackage{graphicx}
\usepackage[utf8]{inputenc} 
\usepackage[T1]{fontenc}    
\usepackage{hyperref}       
\usepackage{url}            
\usepackage{booktabs}       
\usepackage{amsfonts}       
\usepackage{nicefrac}       
\usepackage{microtype}      
\usepackage{xcolor}         
\usepackage{multirow}
\usepackage{amsmath}

\usepackage{pifont}

\newcommand{\xmark}{\ding{55}}%

\begin{document}

\title{Semi-Periodic Activation for Time Series Classification}
%
%
\author{
    José Gilberto Barbosa de Medeiros Júnior, André Guarnier de Mitri \\ Diego Furtado Silva
}
\authorrunning{J. G. B. de Medeiros Júnior et al.}
%
\institute{Universidade de São Paulo, Av. Trab. São Carlense, 400 - Parque Arnold Schimidt, São Carlos - SP, Brazil\\
\email{\{gilberto.barbosa, andremitri\}@usp.br} \\
\email{diegofsilva@icmc.usp.br}}

\maketitle              
\begin{abstract}
This paper investigates the lack of research on activation functions for neural network models in time series tasks. It highlights the need to identify essential properties of these activations to improve their effectiveness in specific domains. To this end, the study comprehensively analyzes properties, such as bounded, monotonic, nonlinearity, and periodicity, for activation in time series neural networks. We propose a new activation that maximizes the coverage of these properties, called LeakySineLU. We empirically evaluate the LeakySineLU against commonly used activations in the literature using 112 benchmark datasets for time series classification, obtaining the best average ranking in all comparative scenarios. 
\end{abstract}


\section{Introduction}

Deep learning models have achieved significant performance improvements across various domains. Consequently, their application in time series tasks has been receiving increasing attention. However, numerous features and techniques remain to be thoroughly explored in this area, leading to the proposal of several new approaches ~\cite{foumani2023deep,wen2023transformers,Wen_2021,ruiz2021mtsc}. Most recent research on deep learning for time series focuses on proposing and evaluating architectures, often adapting those designed for computer vision, such as ResNet ~\cite{wang2016time} and InceptionTime ~\cite{fawaz2020inception}. Other variations aim to extend these architectures beyond classification tasks ~\cite{YANG2022108606,jiang2019resnet}. While these architectures have achieved satisfactory results in their respective domains, they still have gaps to explore.

Deep neural networks can learn effective representations of high-dimensional data, including time series. However, they can suffer from overfitting and other issues that lead to inaccurate results, especially when dealing with periodic data~\cite{ziyin2020neural}. To effectively model time series data, it is crucial to capture inherent patterns such as trends and seasonality to generate suitable representations for the desired tasks.

A notable gap in the literature on deep learning architectures for time series is the development of activation functions capable of adequately transforming the data to representation for a specific task. Some methods propose the use of periodic activation functions, such as $sin(x)$, $cos(x)$, linear combinations of these functions, or more recently, $sin^2(x) + x$, to map these features ~\cite{ziyin2020neural,parascandolo2017taming,831544,zhumekenov2019fourier}. However, even the most recent functions can get stuck in local minima, face issues such as exploding and vanishing gradients, and lack sufficient non-linearity.

\begin{figure}[!htb]
    \centering
    \includegraphics[width=1\columnwidth]{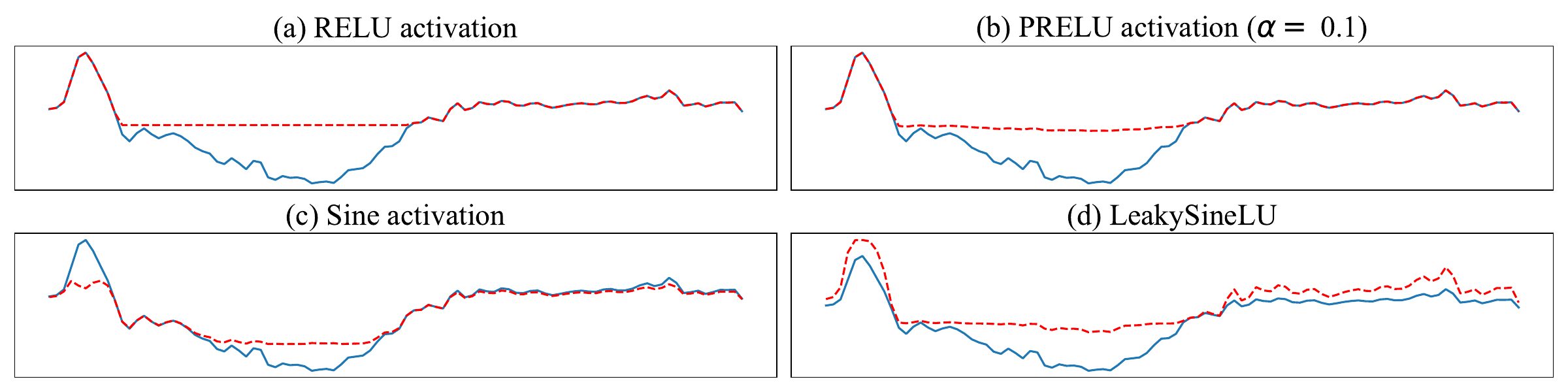}
    \caption{Comparison between different activations. Blue lines represent an input time series, red dashed lines represent the values obtained by applying the activations.}
    \label{fig:activations-comparison}
\end{figure}

Fig.~\ref{fig:activations-comparison} shows the results of activations used in neural networks for time series. ReLU (\textbf{a}) is the most used choice as an activation function in these networks. In the example observed in Fig.~\ref{fig:activations-comparison}, it can not handle features on the input negative values, forcing the network to make a translation operation and put important features on positive values, resulting in a linear activation due to its formula. PReLU (\textbf{b}) comes to solve this problem, as known as dying of ReLU~\cite{kaiming2015activation}, allowing the network to learn a negative slope for the activation by adding a learnable parameter for each neuron. The sine function (\textbf{c}) has been applied on time series problems trying to capture periodic patterns~\cite{sonoda2017activation} once the derivative is also periodic and non-monotonic (the derivative has both positive and negative values), the network can be stuck in local-minimum.

In this work, we propose the LeakySineLU (Fig. \ref{fig:activations-comparison} \textbf{d}), a semi-periodic function as an activation that maintains main properties such as monotonic derivatives, non-linearity, and absence of boundaries. The \textbf{main contributions} of this work include (1) a study of  functions and their basic properties related to time series data; (2) demonstrating that current activation functions can lead to problems when used in time series models, particularly in small architectures; (3) providing a comprehensive evaluation showing that the proposed activation surpasses the use of other functions in two simple and well-known architectures for time series classification.


\section{Definitions and Notations}
\label{sec:definitions}

This section introduces the necessary basic definitions and notations to understand this work.

\begin{definition}\label{def:time-series}
    A time series $\textbf{X}$ is a set of $S$ ordered values $\textbf{X} = (x_1, x_2, \dots, x_S)$ and $x_i \in \mathbb{R}^d$, such that $d$ denotes the number of dimensions of that time series, with $d \in \mathbb{Z}$. Each value $x_i$ is referred here as an \textbf{observation}.
\end{definition}

\begin{definition}\label{def:neural-network}
    A deep neural network, $\mathcal{DNN}$, can be defined as a sequence of functions composed of multiple layers of interconnected neurons. Let $L$ denote the total number of layers in the network, with $l_1, l_2, \dots, l_{{L}}$ representing each layer. A layer $l_i$ consists of $n_i$ neurons, where $i \in 1, \dots, {L}$. Weights and biases associated with that layer $l_i$ are denoted by $\textbf{W}_i$ and $b_i$, respectively. Given an input $\textbf{X}$, the output of a $\mathcal{DNN}$ can be expressed as a sequence of layers applied to that input:

    \begin{equation}
        f(\textbf{X}) = f_\mathcal{L}(f_{\mathcal{L} - 1}(\cdots(f_1(\textbf{X})))),
    \end{equation}

    \noindent where $f_i : \mathbb{R}^{n_{i-1}} \to \mathbb{R}^{n-i}$ represents the function for layer $L_i$. This function can be expressed as:

    \begin{equation}
        f_i(x) = \sigma_i (\mathbf{W}_i x + b_i),
    \end{equation}

    \noindent where $\sigma_i : \mathbb{R}^{n_i} \to \mathbb{R}^{n_i}$ denotes the activation function for the layer $i$, and $x \in \mathbb{R}^{n_{i - 1}}$ represents the input to that layer.
\end{definition}

\begin{definition}\label{def:semi-periodic}
    A semi-periodic activation function is a function where the periodicity is present in its derivative, rather than in the function itself. This means that while the activation function may not be periodically repetitive, its derivative exhibits periodic behavior. Formally, let \(\sigma: \mathbb{R} \to \mathbb{R}\) be an activation function. \(\sigma\) is considered semi-periodic if its derivative \(\sigma'\) satisfies \(\sigma'(x + T) = \sigma'(x)\) for some period \(T > 0\) and for all \(x \in \mathbb{R}\).
\end{definition}

\begin{definition}
    A sub-derivative is a generalization of the concept of a derivative for functions that may not be differentiable in the traditional sense. It extends the idea of a tangent line to non-smooth functions, often used in the context of convex functions and optimization. Formally, let $f: \mathbb{R} \to \mathbb{R}$ be a convex function. A vector $g \in \mathbb{R}$ is called a sub-derivative of $f$ at a point $x_0 \in \mathbb{R}$ if it satisfies the following inequality for all $x \in \mathbb{R}$:

    \begin{equation}
        f(x) \geq f(x_0) + g(x - x_0), 
    \end{equation}

    \noindent the set of all sub-derivatives of $f$ at $x_0$ is called the sub-differential and is denoted by $\partial f(x_0)$. Thus, the sub-differential is given by:

    \begin{equation}
        \partial f(x_0) = \{ g \in \mathbb{R} \mid f(x) \geq f(x_0) + g(x - x_0) \text{ for all } x \in \mathbb{R} \},
    \end{equation}

    \noindent as an example of a sub-derivative we can include the ReLU function $f(x) = \max(0, x)$ the sub-derivative at $x_0 = 0$ is any $g \in [0, 1]$.
\end{definition}


\section{Activation functions}
\label{sec:formulation}

To easily compare the LeakySineLU and the related activation, this section offers a mathematical and theoretical analysis of each activation and their application on time series tasks. In the following subsections, we present the related activations and define the proposed LeakySineLU function, each property, and the comparison between the main activations.

\subsection{Related Activations}

To compare the proposed activation with various approaches, activation functions encompassing one or more specific properties were selected, and in some cases, functions that do not possess certain properties were also included. Table~\ref{tab:activations} presents all related activations compared in this work, their formula and derivatives.

\begin{table}[ht]
    \centering
    \caption{Activation functions compared in this work, their base and derived formulas.}
    \label{tab:activations}
    \begin{tabular}{c|c|c}
         \textbf{Activation} & \textbf{Formula} & \textbf{Derivative} \\
         \hline\hline
         Sigmoid & $1/(1 + e^{-x})$ & $\sigma(x)(1 - \sigma(x))$ \\
         TanH & $(e^x - e^{-x})/(e^x + e^{-x})$ & $1 - \tanh^2(x)$ \\
         Sine & $\sin(x)$ & $\cos(x)$ \\
         ReLU & $\max(0, x)$ & $\{ 0, 1 \}$ \\
         ELU & $\max(\alpha (e^x - 1),x)$ & $\{ \alpha e^x, 1 \}$ \\
         PReLU & $\max(\alpha x, x)$ & $\{ \alpha, 1 \}$ \\
         GeLU & $x \Phi(x)$ & $(\alpha x)/(\sqrt{2\pi})e^{-\frac{(\alpha x)^2}{2}} + \Phi(\alpha x) $ \\
         SiLU & $x \sigma(x)$ & $\sigma(x)(1 + x (1 - \sigma(x)))$ \\
         Snake & $x + \sin^2 (ax)/a$ & $1 + \sin(2ax)$ \\
         \hline
    \end{tabular}
\end{table}

This selection includes functions with upper and lower bounds, such as the sigmoid and hyperbolic tangent. To represent the family of periodic functions, the sine function was chosen. To cover functions derived from the ReLU family, the following were selected: ReLU, ELU as an exponential linear unit (PReLU) for its learnable parameters, GeLU for its continuity over the entire domain, and SiLU for combining the sigmoid strategy with a linear unit. At last, Snake was chosen due to its approach being the most similar to our proposal.

\subsection{LeakySineLU}

In this work, we propose the LeakySineLU, defined by the following equation:

\begin{equation}
\label{eq:leakysinelu}
    \sigma(x) = \begin{cases}
        \sin^{2}(x) + x & \text{if} ~ x > 0, \\
        \frac{\sin^{2}(x) + x}{2} & \text{otherwise}.
    \end{cases}
\end{equation}

The derivative of this equation can be defined as:

\begin{equation}
\label{eq:derivative_leakysinelu}
    \sigma'(x) = \begin{cases}
        \sin(2x) + 1 & \text{if} ~ x > 0, \\
        \frac{\sin(2x) + 1}{2} & \text{otherwise}.
    \end{cases}
\end{equation}

\begin{figure}[htb]
    \centering
    \includegraphics[width=0.7\textwidth]{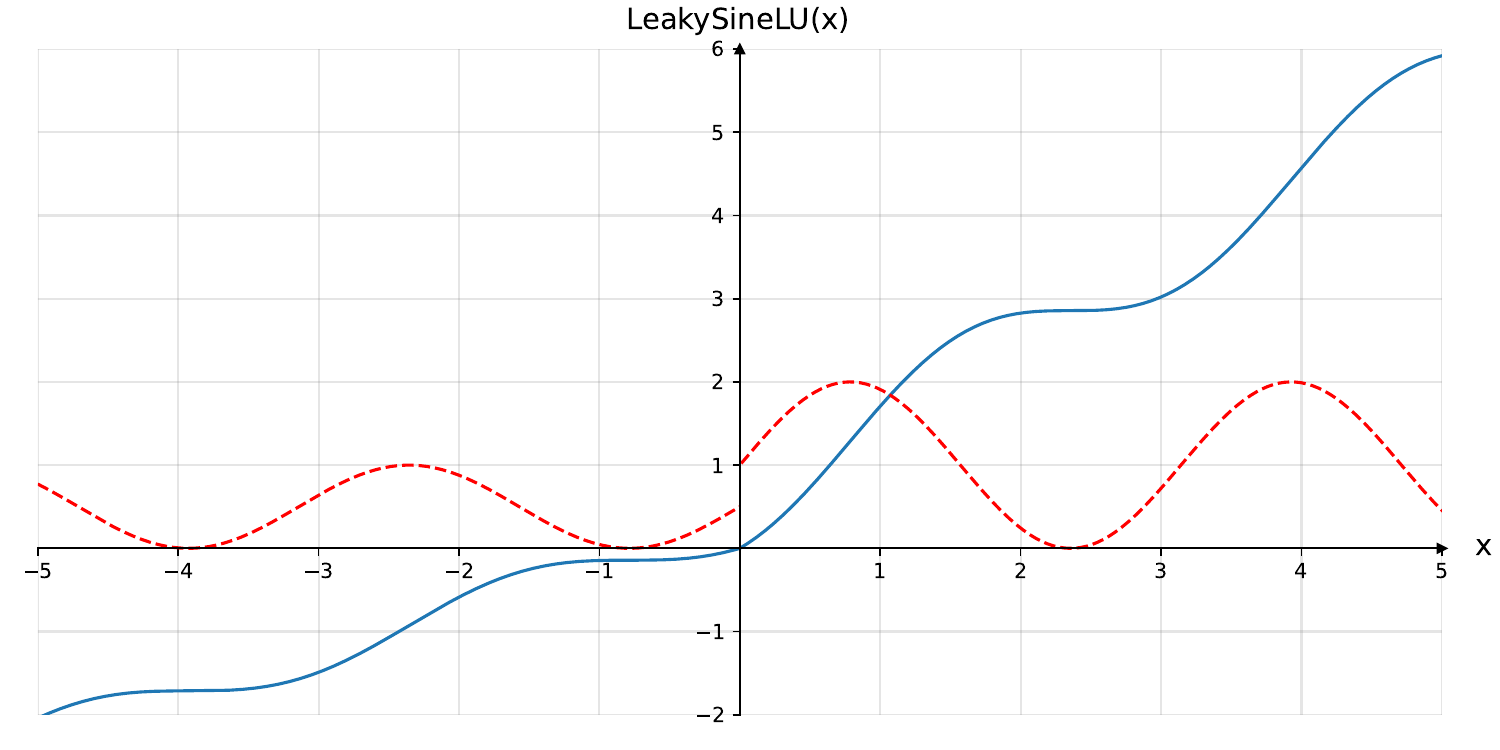}
    \caption{LeakySineLU defined in Equation~\ref{eq:leakysinelu} represented in blue, and its derivative defined in Equation~\ref{eq:derivative_leakysinelu} in red.}
    \label{fig:leakysinelu}
\end{figure}

Figure~\ref{fig:leakysinelu} presents the image domain and the derivative of LeakySineLU. It's possible to note that the derivate is not continuous when $x = 0$. As in ReLU and PReLU activations, this problem is solved using sub-derivatives and sub-gradients (c.f. Section~\ref{sec:diffentiability}). However, we maintain a periodic behavior on the derivative in both domains, negative and positive, and an increased non-linearity by the difference on the scale for these domains.

\subsection{Boundness}

Over the years, many studies have shown that bounded activation functions, such as hyperbolic tangent or sigmoid, could reach excellent results in many different tasks. Although these results are specifically reported when using shallow network architectures~\cite{pedamonti2018comparison,liew2016bounded}. However, training neural networks that use this type of function with time series data may cause a loss of information, which occurs when the series contains values outside the boundaries, causing the problem of vanishing gradients~\cite{bengio1994gradient}.

Neural networks that use unbounded activation functions, with regards to a non-polynomial activation, are universal approximators and seem to attenuate the vanishing gradient problem~\cite{sonoda2017activation,pinkus1999activation,nair2010rectified}. However, activations that are upper- or lower-bounded, such as the lower-bounded ReLU, can cause a loss of information on negative observations of the time series, which results in a problem known as ``dying ReLU'' \cite{kaiming2015activation}. Other activations such as LeakyReLU, ELU~\cite{djork2016elu}, SiLU~\cite{stefan2017silu}, PReLU, and GeLU~\cite{hendrycks2016gelu} have been proposed to address this issue, either using a negative slope different from $0$ or specific weight initializations. But except for LeakyReLU and PReLU, all others have lower bounds, even if different from $0$.

\begin{figure}[t]
    \centering
    \includegraphics[width=1\textwidth]{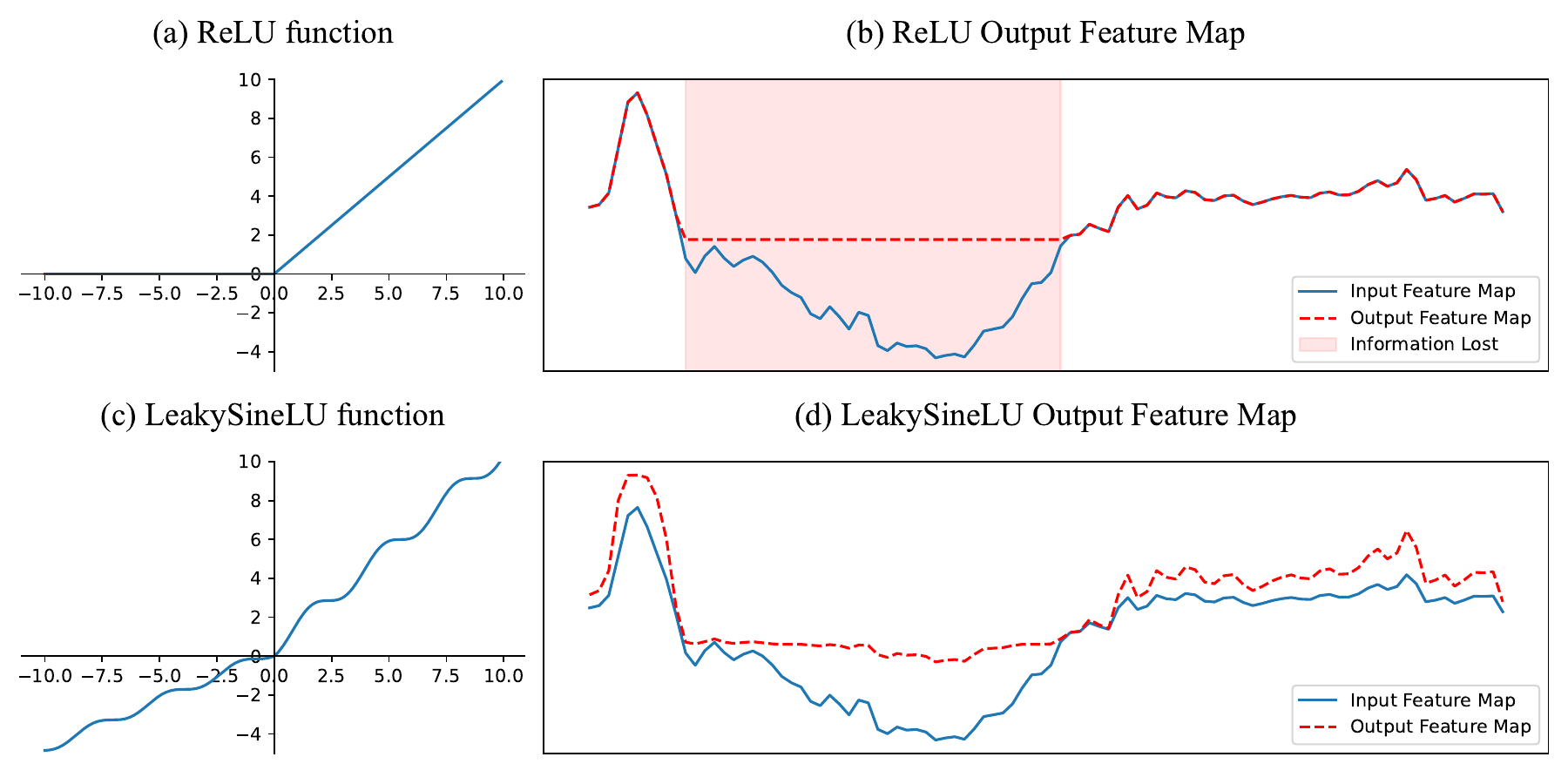}
    \caption{Effects of boundaries on time series feature maps. Red region presents the information lost, which means that $f(x) = 0$ for that values.}
    \label{fig:activation-bounds-ts}
\end{figure}

Figure~\ref{fig:activation-bounds-ts} presents a comparison between ReLU and LeakySineLU functions applied to a time series. Note that in this case, for negative values of the time series, we may lose some information, which means that $f(x) = 0$. The region of the time series that this occurs is represented by a red highlighted background on that series. This can result in the previously cited dying of the ReLU problem in those neurons. LeakySineLU maintains a negative slope while preserving the negative-values of the original series. An argument in favor of ReLU or its derivations is that we can maintain control of the value range for that series by applying some normalization method~\cite{lima2023comparetime}, but with this, we put all values on the positive image domain, which means that we have a linear activation and with that comes the lack of non-linearity that will be explained in Section~\ref{sec:nonlinearity}.

To analyze the boundness of LeakySineLU activation function we rely on the limits of $f(x) \to +\infty$ and $f(x) \to -\infty$. If one of these limits is equals to a constant value, then the function is upper or lower bounded.

\begin{equation}
    \lim_{x \to +\infty} \sigma(x) = \lim_{x \to +\infty} \sin^{2}(x) + x = \sin^{2}(+\infty) + \infty = +\infty,
\end{equation}

\begin{equation}
    \lim_{x \to -\infty} \sigma(x) = \lim_{x \to -\infty} \frac{\sin^{2}(x) + x}{2} = \frac{\sin^{2}(- \infty) - \infty}{2} = -\infty,
\end{equation}

\noindent as we can see, both of the limits are going to $+ \infty$ and $- \infty$, with this we can conclude that the LeakySineLU have no boundaries. Table ~\ref{tab:activations_bounds} present a comparison between each related activation functions and their boundaries.

\begin{table}[t]
    \centering
    \caption{Negative and positive limits as $x$ tends to $+\infty$ or $-\infty$ for each of the activations.}
    \label{tab:activations_bounds}
    \begin{tabular}{c|cccccccccc}
          & \textbf{Sigmoid} & \textbf{TanH} & \textbf{Sine} & \textbf{ReLU} & \textbf{ELU} & \textbf{PReLU} & \textbf{GeLU} & \textbf{SiLU} & \textbf{Snake} & \textbf{Ours} \\
          \hline\hline
         \textbf{Lower} & 0 & -1 & 0 & 0 & -1 & $-\infty$ & 0 & 0 & $-\infty$ & $-\infty$ \\
         \textbf{Upper} & 1 & 1 & 1 & $+\infty$ & $+\infty$ & $+\infty$ & $+\infty$ & $+\infty$ & $+\infty$ & $+\infty$ \\
         \hline
    \end{tabular}
\end{table}

\subsection{Non-linearity}\label{sec:nonlinearity}

Non-linearity is a critical point for activation functions~\cite{bengio1994gradient}. All activations want to increase the non-linearity present in neural networks. Activations such as ReLU, GeLU, ELU, PReLU, SiLU present this property. Nevertheless, when we isolate the positive side of this activations, its behavior in the domain is linear, which can hinder learning for some tasks, such as when we normalize the entire series to only have positive values. Consider a basic neuron defined before:

\begin{equation*}
    f_i(x) = \sigma_i (\mathbf{W}_i x + b_i),
\end{equation*}

\noindent where  $\mathbf{W}_{i}$ and $b_i$ are the weight matrix and bias vector of the $i^{th}$ layer, and $f_i(x)$ is the output of the previous layer. We can represent the output of the neural network as:
    
\begin{equation}
    f(\mathbf{X}) = \mathbf{W}_{L} \sigma(\mathbf{W}_{L-1} \sigma(\cdots \sigma( \mathbf{W}_{2}\sigma(\mathbf{W}_{1} {\mathbf{X}} + {b}_{1}) + {b}_{2}) + {b}_{L-1} ) + {b}_{L},
\end{equation}
    
\noindent where  $\mathbf{X}$ is the input time series to the neural network. If the neural network uses only linear activations, such as only the positive side of ReLU, the $\sigma$ can be ignored.

\begin{equation}
    f(\mathbf{X}) = \mathbf{W}_{L} (\mathbf{W}_{L-1} (\cdots ( \mathbf{W}_{2} (\mathbf{W}_{1} {\mathbf{X}} + {b}_{1}) + {b}_{2}) + {b}_{L-1} ) + {b}_{L},
\end{equation}

We can now denote a total weight and bias by respectively:

\begin{equation}
    W_{\mathcal{DNN}} = \mathbf{W}_L \mathbf{W}_{L-1} \cdots \mathbf{W}_1,
\end{equation}

\begin{equation}
    b_{\mathcal{DNN}} = \mathbf{W}_L \mathbf{W}_{L-1} \cdots \mathbf{W}_2 b_1 + \mathbf{W}_L \mathbf{W}_{L-1} \cdots \mathbf{W}_3 b_2 + \cdots + \mathbf{W}_L b_{L-1} + b_L,
\end{equation}

\noindent with this two definitions we can simplify the equation of a deep neural network that uses a linear activation with the following equation:

\begin{equation}
    f(\mathbf{X}) = \mathbf{W}_{\mathcal{DNN}} \mathbf{X} + b_{\mathcal{DNN}},
\end{equation}

\noindent note that this equation have a form of a linear model.

    
    

\subsection{Differentiability}\label{sec:diffentiability}

The differentiability of activation functions is critically important in neural networks, as it facilitates the application of gradient-based optimization methods, such as the backpropagation algorithm. These methods require the computation of derivatives (or gradients) of the activation functions concerning their parameters to adjust the network's weights efficiently. Differentiable activation functions ensure the smooth propagation of gradients back through the layers of the neural network, which is essential for effective minimization of the loss function~\cite{rasamoelina2020activation_review,kaiming2015activation}.

As shown in Equation~\ref{eq:derivative_leakysinelu}, the derivative for LeakySineLU can be calculated in two parts. Figure~\ref{fig:leakysinelu} presents the proposed activation and its derivative. Note that there is a discontinuity when $x = 0$. By using the definition of sub-derivatives we can define that the derivatives at $x = 0$ are given by:

\begin{equation*}
    \partial f(x_0) = \{ \sin(x_0) + 1, (\sin(x_0) + 1)/2 \} = \{ 1, 0.5 \},
\end{equation*}

\noindent that way we can use gradient-based optimization methods and the LeakySineLU function in neural networks. This is the same strategy already used for functions such as ReLU and PReLU.

\subsection{Periodicity}

The motivation behind the periodicity is based on the Universal Extrapolation Theorem~\cite{ziyin2020neural} and the generalized Fourier Series. Consider the definition of this general form of Fourier Series:

\begin{equation}
    x(t) = \frac{a_0}{2} + \sum^\infty_{n=1} a_n \cos(2\pi \frac{nt}{T}) + b_n \sin(2\pi \frac{nt}{T}),
\end{equation}

\noindent where  $a_0$, $a_n$, $b_n$ are constants and $T$ is a elemental period. Let $a_0$ be an array in a $\mathbb{R}^d$ space, the result of a scalar division $a_0/2$ is a operation that maps $f: \mathbb{R}^d \to \mathbb{R}^d$. We can rename the result of $a_0/2$ to a bias vector and the constants $a_n$ and $b_n$ by weights $w_{0n}$ and $w_{1n}$, that way the formula will be:

\begin{equation}
    x(t) = b + \sum^\infty_{n=1} w_{0n} \cos(2\pi\frac{nt}{T}) + w_{1n} \sin(2\pi\frac{nt}{T}),
\end{equation}

\noindent now considering the mathematical equivalence of $\sin$ and $\cos$, $\sin(x+3\pi/2) = \cos(x)$, and considering each input for $\sin$ and $\cos$ as $x_0$ and $x_1$ we can rewrite the formula as:

\begin{equation}
    x(t) = b + \sum^\infty_{n=1} w_{0n} \sin(x_0) + w_{1n} \sin(x_1)
\end{equation}

\noindent that way we can group $w_{0n}$ and $w_{1n}$ is a weight matrix $\mathbf{W}$, and $x_0$ and $x_1$ also in a input matrix $\mathbf{X}$. The $\sin$ will operate over a matrix and return an output in the same dimension, by the property of matrix multiplication, we can arrange the columns and rows to perform the multiplication over the correct values. Resulting in a final formula similar to a neural network layer with $\sin$ as activation:

\begin{equation}
    x(t) = b_{ias} + \sum^\infty_{n=1} W_n \sin(X)
\end{equation}

\subsection{Monotonic}

A monotonic activation function is a function that only increases or decreases as its input increases, being allowed to stay constant at some ranges. Sigmoid, ReLU, and hyperbolic tangent are examples of functions of this type.

The monotonicity of an activation function is a common property found in the literature, although it is not mandatory. For instance, Mish~\cite{misra2020mish} and GeLU~\cite{hendrycks2016gelu} are exceptions to this property, allowing the network to capture more complex nonlinear relationships between data while maintaining differentiability across its entire domain. However, the choice of non-monotonicity of Mish and GeLU may restrict its use in some tasks. \cite{sopena1999monotonic} has shown that periodic activation functions are a promising alternative to monotonic activation functions in neural networks. At the time of that work, no studies demonstrated that functions such as sine could fall into local minima~\cite{parascandolo2017taming}.

The neural network can use monotonic activation to learn a mapping function that is easier to optimize since the derivative only increases or decreases in one direction. That way, the gradient descendent algorithms can work more efficiently.

\begin{table}[ht]
    \centering
    \caption{Comparison between activation functions and their monotonic property.}
    \label{tab:activation_properties}
    \begin{tabular}{cccccccccc}
          \textbf{Sigmoid} & \textbf{TanH} & \textbf{Sine} & \textbf{ReLU} & \textbf{ELU} & \textbf{PReLU} & \textbf{GeLU} & \textbf{SiLU} & \textbf{Snake} & \textbf{Ours} \\
          \hline\hline
          \checkmark & \checkmark & \xmark & \checkmark & \checkmark & \checkmark & \xmark & \xmark & \checkmark & \checkmark \\
         \hline
    \end{tabular}
\end{table}

\section{Experiments}

This section assesses the LeakySineLU in different experiments to evaluate its applicability in classification tasks involving time series. Initially, the experiments were executed and discussed in 112 equal-length subset\footnote{All datasets that does not have "Vary" as time series length in the repository.} datasets from the UCR repository~\cite{UCRArchive2018} (experiments are available in the project repository\footnote{\url{https://github.com/jose-gilberto/leakysinelu}}). We conduct the experiments in two well-established architectures. The first is a Multi-Layer Perceptron (MLP)~\cite{wang2016time,fawaz2019tsreview}, representing the most basic and straightforward architectures. The second choice is a Fully Convolutional Network (FCN)~\cite{wang2016time,fawaz2019tsreview}, an architecture representing networks that use convolutional structures to obtain more accurate results.

MLP consists of a network with two blocks, each composed of a dropout ($p=0.1$ and $p=0.2$, respectively, in each block) and a layer with 500 neurons. Next, a last block is used, composed of a dropout with $p=0.3$ and another layer with the number of classes being the number of neurons, where a softmax activation is applied for multi-class classification or just $1$ neurons for binary classification. Each layer includes the activation applied to that experiment. FCN comprises three convolutional blocks (128, 256, and 128 channels, respectively) with kernel sizes of $8$, $5$, and $3$ applied in each layer. The output of these blocks is used in an average pooling and then passed to a layer with the number of neurons equal to the number of classes for multi-class problems or one neuron for binary classification. A softmax activation is used in cases with more than one output neuron. For convolutions, the same padding is used to preserve the input dimensions, no max pooling is applied to the series, and the stride parameter is always equal to 1. The activation used in each block varies according to the experiment.

\subsection{Experimental Setup}

We conducted experiments using the two chosen architectures, MLP and FCN, to evaluate the applicability of the proposed activation function in time series classification tasks. The MLP network was trained using the Adadelta optimizer \cite{zeiler2012adadelta}, with a learning rate of $1.0$ and $1000$ training epochs. The FCN network was trained using the Adam optimizer \cite{kingma2017adam}, with a learning rate of 0.001 and 2000 training epochs. The loss function used in the experiments was the cross-entropy for multi-class classification and binary cross-entropy for binary classification tasks. The performance of the models was evaluated in terms of accuracy.


\section{Results}

\begin{figure}
    \centering
    \includegraphics[width=1\textwidth]{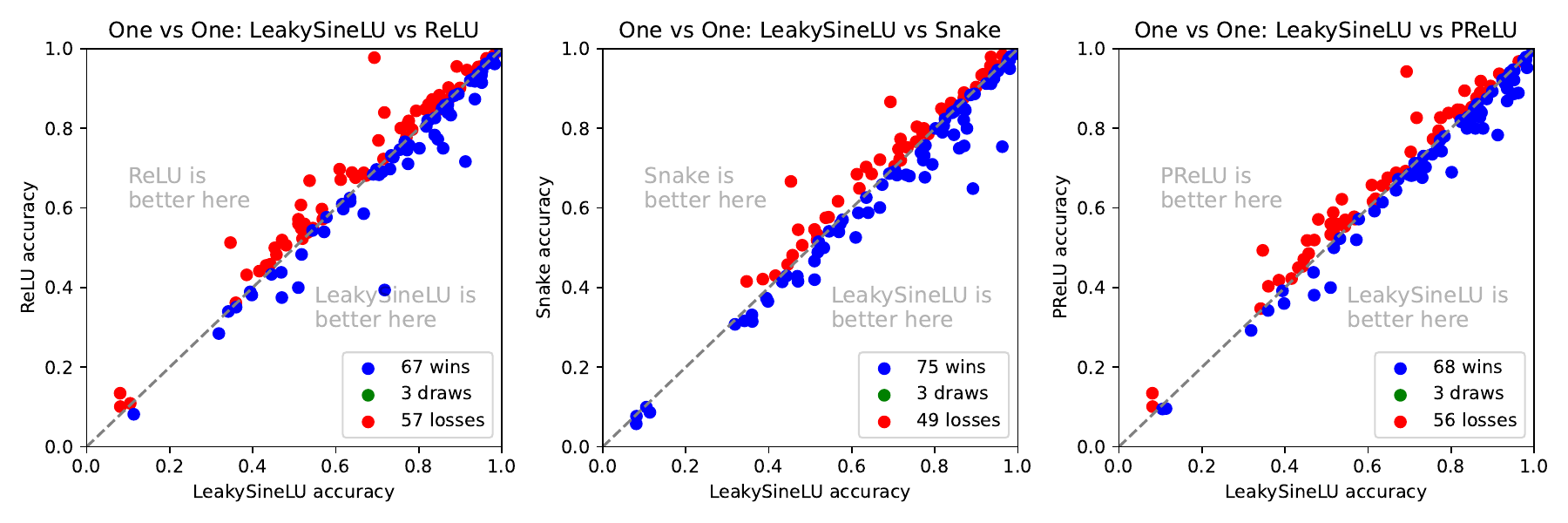}
    \caption{Comparison between LeakySineLU, ReLU, PReLU, and Snake on an MLP network among the 112 equal-length datasets.}
    \label{fig:one-vs-one}
\end{figure}

This section discusses the results of LeakySineLU and all the other approaches over the 112 equal-length UCR datasets. Following~\cite{fawaz2019tsreview}, we use the mean accuracy averaged over the experiments on the test set. In Fig.~\ref{fig:one-vs-one}, we present a one-versus-one comparison between the LeakySineLU function, ReLU, Snake, and PReLU in a MLP network. We provide these comparisons to analyze the performance with the most used activation in time series tasks (ReLU), the most similar activation (Snake), and a robust alternative to ReLU (PReLU).

Each point in these plots represents one dataset. The x-axis represents the accuracy of the test set using classifier-x, and the y-axis represents the ones using classifier-y. The diagonal line shows a draw-line between these models. Note that the LeakySineLU improves the accuracy compared to the other activations, besides the competitive results compared to the ReLU and PReLU.

\begin{figure}[ht]
    \centering
    \includegraphics[width=0.8\textwidth
    ]{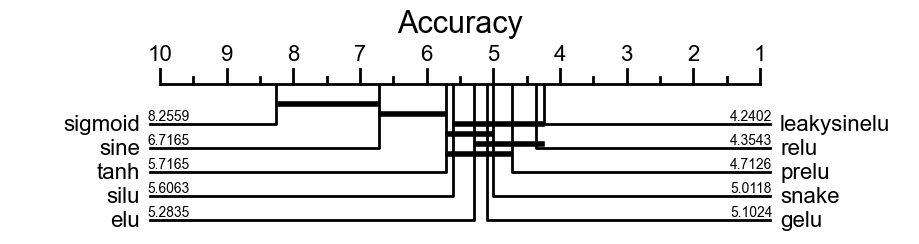}
    \caption{Critical Difference diagram with the average rank of each classifier over the 112 equal-length datasets of the UCR archive. The horizontal black bar denotes that the models connected with it have no statistical difference.}
    \label{fig:cdd-mlp}
\end{figure}

As~\cite{fawaz2019tsreview}, we based our analysis on the recommendation of using the Friedman test to reject the null hypothesis and perform a pairwise post-hoc analysis with the average ranking comparison replaced by the Wilcoxon signed-rank test with Holm's alpha (5\%) correction. To facilitate the visualization of the rankings and the comparison between the methods, we rely on the critical difference diagram. Fig.~\ref{fig:cdd-mlp} presents the Critical Difference diagram generated over the experiments in the UCR repository with the MLP network. Note that the LeakySineLU function is in the $1^{\text{st}}$ position with an average ranking of $4.2402$, besides not having a statistical difference from Snake and ReLU derivations.

\begin{figure}[ht]
    \centering
    \includegraphics[width=1\textwidth]{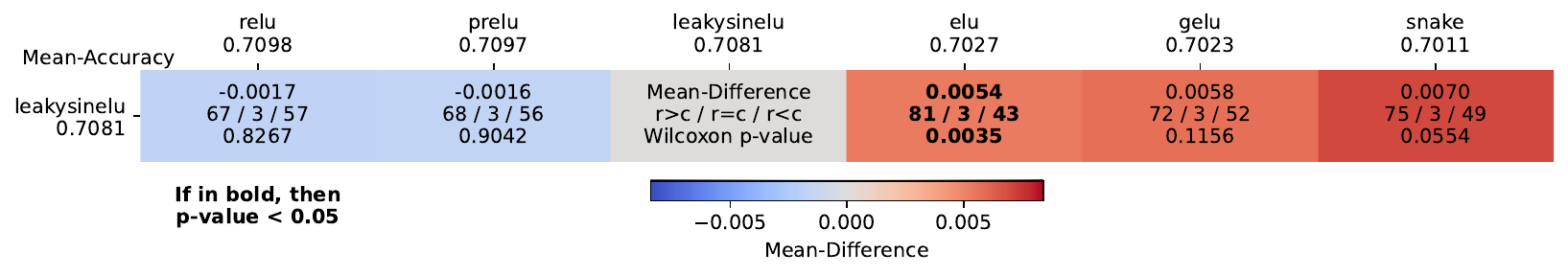}
    \caption{Multi-Comparison Matrix of LeakySineLU in an MLP compared to the other statistical equivalent approaches in Fig.~\ref{fig:cdd-mlp}. \textbf{Bold indicates p-value $\leq$ 0.05}.}
    \label{fig:mcm-comparison-mlp-1}
\end{figure}

To continue the analysis of the results, we rely on the Multi-Comparison Matrix (MCM)~\cite{ismail2023approach}. The MCM is a more stable tool for evaluating methods in terms of variation in the models used for the comparison. Fig.~\ref{fig:mcm-comparison-mlp-1} presents the MCM that compares the LeakySineLU in an MLP network with the other activations that do not show a statistical difference in Fig.~\ref{fig:cdd-mlp}. This MCM uses the average accuracy on the UCR equal-length datasets as an ordering metric. Observe that, besides the statistical equivalence between these activations, the LeakySineLU wins in the majority of the datasets when compared to each other statistically equivalent activation.

\begin{figure}
    \centering
    \includegraphics[width=1\textwidth]{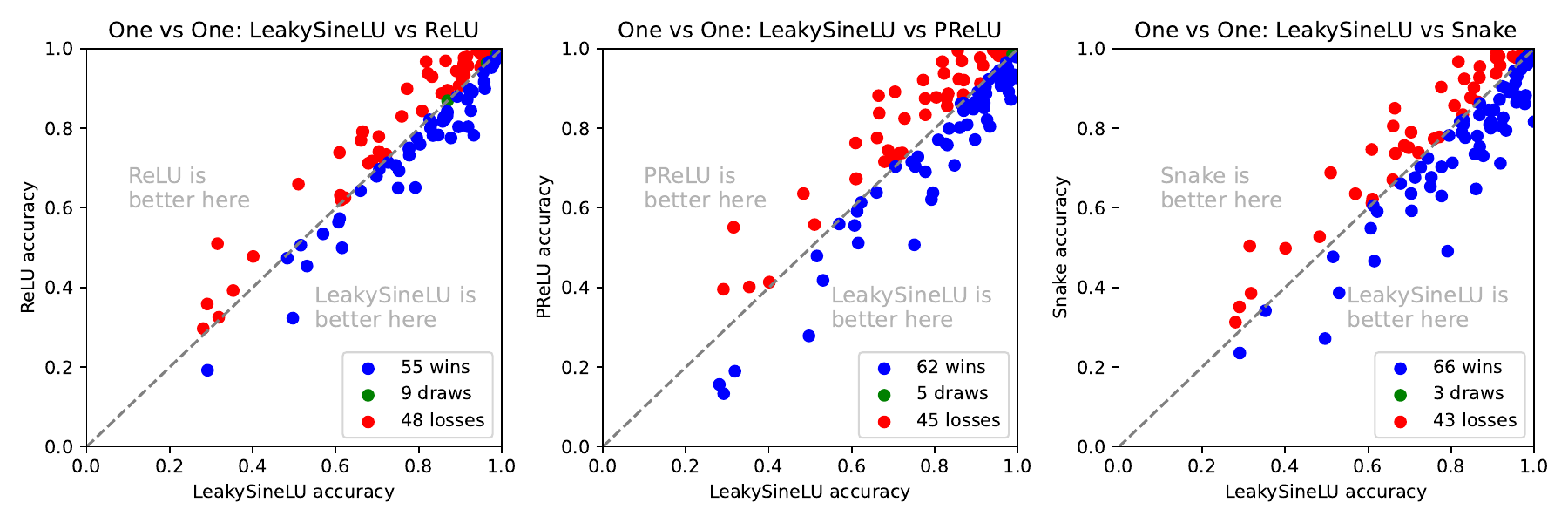}
    \caption{Comparison between LeakySineLU, ReLU, PReLU, and Snake on an FCN network among the 112 equal-length datasets.}
    \label{fig:one-vs-one-fcn}
\end{figure}

Moving on to the FCN analysis, we used the same activations compared previously during the experiments. However, for simplicity reasons, we analyzed in more depth the activations that have already shown to be statistically equivalent, namely LeakySineLU, ReLU, PReLU, ELU, GeLU, and Snake. Fig.~\ref{fig:one-vs-one-fcn} presents a one-versus-one comparison between the LeakySineLU function, ReLU, Snake, and PReLU in a FCN network. In this case, the number of victories remains in favor of LeakySineLU. It is possible to observe that the network, even using convolution filters, still benefits from an activation designed and focused on time series problems.

\begin{figure}[ht]
    \centering
    \includegraphics[width=0.8\textwidth
    ]{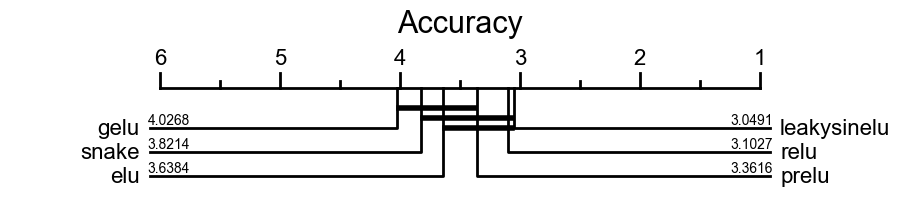}
    \caption{Critical Difference diagram with the average rank of each FCN classifier over the 112 equal length datasets of the UCR archive.}
    \label{fig:cdd-fcn}
\end{figure}

Fig.~\ref{fig:cdd-fcn} presents the Critical Difference diagram generated over the experiments in the UCR repository with the FCN network. In this case, the critical difference diagram shows that LeakySineLU is statistically different from ELU. However, it is worth remembering that, as highlighted in~\cite{ismail2023approach}, an average ranking can be prone to errors. To solve this, we also used MCM to analyze the FCN results. However, it is still worth noting that LeakySineLU is the best-ranked activation function, with an average ranking of $3.0491$. When we compare its performance with ReLU (second best-ranked), the proposed function wins in most cases.

\begin{figure}[ht]
    \centering
    \includegraphics[width=1\textwidth]{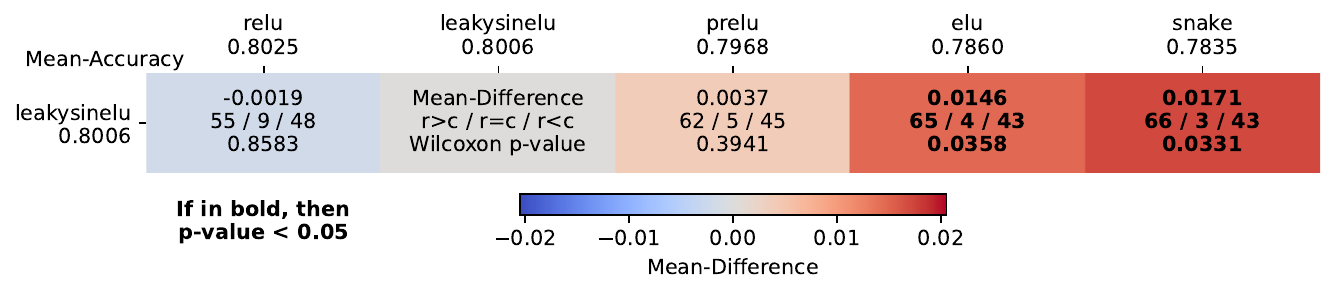}
    \caption{Multi-Comparison Matrix of LeakySineLU in a FCN compared to the other statistical equivalent approaches in Fig.~\ref{fig:cdd-fcn}. \textbf{Bold indicates p-value $\leq$ 0.05}.}
    \label{fig:mcm-comparison-fcn-1}
\end{figure}

Fig.~\ref{fig:mcm-comparison-fcn-1} presents the MCM that compares the LeakySineLU in an FCN network. As mentioned, when varying the analysis to the average accuracy and a statistical test recommended in the literature, MCM provides us with more detailed information. It is possible to notice two things, the first of which is that LeakySineLU is statistically different from ELU and Snake in this analysis. The second is that despite being equivalent to ReLU and PReLU, it can be noted that LeakySineLU still wins the majority of cases when compared to these two activations, with the number of wins being 55 and 62, respectively.

\section{Conclusion}

In this work, we studied and identified meaningful potential harmful properties of neural network activations for time series data. Our study encompasses mathematical analysis of these properties as well as their study for tasks related to time series. We identified weaknesses in the main activations in the literature and, based on this study, we proposed the LeakySineLU, an activation that overcomes varied limitations of activation functions commonly used on neural network architectures for time series tasks. Through an empirical study, we conducted a series of experiments on benchmark datasets from the literature, showing that the proposed activation may be a relevant alternative to be considered for use in time series classification tasks. Future works include improvement in well-defined architectures for time series classification such as InceptionTime~\cite{fawaz2020inception} and ResNet~\cite{wang2016time} using the proposed activation. The impact of activation choices in other time series-related tasks such as extrinsic regression and forecasting exploring the use of LeakySineLU and other periodic properties.

\section*{Acknowledgement}

The authors would like to acknowledge the financial support by grants \#2022/03176-1, \#2023/02680-0, \#2023/11775-5, \#2023/05041-9 and grant \#2023/11745-9, São Paulo Research Foundation (FAPESP). We would also like to thank the creators of the UCR data repository for making the time series datasets publicly available.

%
%

\bibliographystyle{splncs04}
\bibliography{references}

\begin{thebibliography}{10}
\providecommand{\url}[1]{\texttt{#1}}
\providecommand{\urlprefix}{URL }
\providecommand{\doi}[1]{https://doi.org/#1}

\bibitem{bengio1994gradient}
Bengio, Y., Simard, P., Frasconi, P.: Learning long-term dependencies with gradient descent is difficult. IEEE Transactions on Neural Networks  \textbf{5}(2),  157--166 (1994). \doi{10.1109/72.279181}

\bibitem{djork2016elu}
Clevert, D., Unterthiner, T., Hochreiter, S.: Fast and accurate deep network learning by exponential linear units (elus). In: Bengio, Y., LeCun, Y. (eds.) 4th International Conference on Learning Representations, {ICLR} 2016, San Juan, Puerto Rico, May 2-4, 2016, Conference Track Proceedings (2016), \url{http://arxiv.org/abs/1511.07289}

\bibitem{UCRArchive2018}
Dau, H.A., Keogh, E., Kamgar, K., Yeh, C.C.M., Zhu, Y., Gharghabi, S., Ratanamahatana, C.A., Yanping, Hu, B., Begum, N., Bagnall, A., Mueen, A., Batista, G., Hexagon-ML: The ucr time series classification archive (October 2018), \url{https://www.cs.ucr.edu/~eamonn/time_series_data_2018/}

\bibitem{stefan2017silu}
Elfwing, S., Uchibe, E., Doya, K.: Sigmoid-weighted linear units for neural network function approximation in reinforcement learning. CoRR  \textbf{abs/1702.03118} (2017), \url{http://arxiv.org/abs/1702.03118}

\bibitem{fawaz2019tsreview}
Fawaz, H.I., Forestier, G., Weber, J., Idoumghar, L., Muller, P.A.: Deep learning for time series classification: a review. Data Mining and Knowledge Discovery  \textbf{33}(4),  917--963 (mar 2019). \doi{10.1007/s10618-019-00619-1}, \url{https://doi.org/10.1007\%2Fs10618-019-00619-1}

\bibitem{fawaz2020inception}
Fawaz, H.I., Lucas, B., Forestier, G., Pelletier, C., Schmidt, D.F., Weber, J., Webb, G.I., Idoumghar, L., Muller, P.A., Petitjean, F.: {InceptionTime}: Finding {AlexNet} for time series classification. Data Mining and Knowledge Discovery  \textbf{34}(6),  1936--1962 (sep 2020). \doi{10.1007/s10618-020-00710-y}, \url{https://doi.org/10.1007\%2Fs10618-020-00710-y}

\bibitem{foumani2023deep}
Foumani, N.M., Miller, L., Tan, C.W., Webb, G.I., Forestier, G., Salehi, M.: Deep learning for time series classification and extrinsic regression: A current survey (2023)

\bibitem{kaiming2015activation}
He, K., Zhang, X., Ren, S., Sun, J.: Delving deep into rectifiers: Surpassing human-level performance on imagenet classification. In: 2015 IEEE International Conference on Computer Vision (ICCV). pp. 1026--1034 (2015). \doi{10.1109/ICCV.2015.123}

\bibitem{hendrycks2016gelu}
Hendrycks, D., Gimpel, K.: Bridging nonlinearities and stochastic regularizers with gaussian error linear units. CoRR  \textbf{abs/1606.08415} (2016), \url{http://arxiv.org/abs/1606.08415}

\bibitem{ismail2023approach}
Ismail-Fawaz, A., Dempster, A., Tan, C.W., Herrmann, M., Miller, L., Schmidt, D.F., Berretti, S., Weber, J., Devanne, M., Forestier, G., et~al.: An approach to multiple comparison benchmark evaluations that is stable under manipulation of the comparate set. arXiv preprint arXiv:2305.11921  (2023)

\bibitem{kingma2017adam}
Kingma, D.P., Ba, J.: Adam: A method for stochastic optimization (2017)

\bibitem{liew2016bounded}
Liew, S.S., Khalil-Hani, M., Bakhteri, R.: Bounded activation functions for enhanced training stability of deep neural networks on visual pattern recognition problems. Neurocomputing  \textbf{216},  718--734 (2016). \doi{https://doi.org/10.1016/j.neucom.2016.08.037}, \url{https://www.sciencedirect.com/science/article/pii/S0925231216308797}

\bibitem{lima2023comparetime}
Lima, F.T., Souza, V.M.: A large comparison of normalization methods on time series. Big Data Research  \textbf{34},  100407 (2023). \doi{https://doi.org/10.1016/j.bdr.2023.100407}, \url{https://www.sciencedirect.com/science/article/pii/S2214579623000400}

\bibitem{misra2020mish}
Misra, D.: Mish: A self regularized non-monotonic activation function (2020)

\bibitem{nair2010rectified}
Nair, V., Hinton, G.E.: Rectified linear units improve restricted boltzmann machines. In: Proceedings of the 27th international conference on machine learning (ICML-10). pp. 807--814 (2010)

\bibitem{parascandolo2017taming}
Parascandolo, G., Huttunen, H., Virtanen, T.: Taming the waves: sine as activation function in deep neural networks (2017), \url{https://openreview.net/forum?id=Sks3zF9eg}

\bibitem{pedamonti2018comparison}
Pedamonti, D.: Comparison of non-linear activation functions for deep neural networks on mnist classification task (2018)

\bibitem{pinkus1999activation}
Pinkus, A.: Approximation theory of the mlp model in neural networks. Acta Numerica  \textbf{8},  143–195 (1999). \doi{10.1017/S0962492900002919}

\bibitem{rasamoelina2020activation_review}
Rasamoelina, A.D., Adjailia, F., Sinčák, P.: A review of activation function for artificial neural network. In: 2020 IEEE 18th World Symposium on Applied Machine Intelligence and Informatics (SAMI). pp. 281--286 (2020). \doi{10.1109/SAMI48414.2020.9108717}

\bibitem{ruiz2021mtsc}
Ruiz, A.P., Flynn, M., Large, J., Middlehurst, M., Bagnall, A.: The great multivariate time series classification bake off: a review and experimental evaluation of recent algorithmic advances. Data Mining and Knowledge Discovery  \textbf{35},  401--449 (2021). \doi{https://doi.org/10.1007/s10618-020-00727-3}

\bibitem{831544}
Silvescu, A.: Fourier neural networks. In: IJCNN'99. International Joint Conference on Neural Networks. Proceedings (Cat. No.99CH36339). vol.~1, pp. 488--491 vol.1 (1999). \doi{10.1109/IJCNN.1999.831544}

\bibitem{sonoda2017activation}
Sonoda, S., Murata, N.: Neural network with unbounded activation functions is universal approximator. Applied and Computational Harmonic Analysis  \textbf{43}(2),  233--268 (2017). \doi{https://doi.org/10.1016/j.acha.2015.12.005}, \url{https://www.sciencedirect.com/science/article/pii/S1063520315001748}

\bibitem{sopena1999monotonic}
Sopena, J., Romero, E., Alquezar, R.: Neural networks with periodic and monotonic activation functions: a comparative study in classification problems. In: 1999 Ninth International Conference on Artificial Neural Networks ICANN 99. (Conf. Publ. No. 470). vol.~1, pp. 323--328 vol.1 (1999). \doi{10.1049/cp:19991129}

\bibitem{wang2016time}
Wang, Z., Yan, W., Oates, T.: Time series classification from scratch with deep neural networks: A strong baseline (2016)

\bibitem{Wen_2021}
Wen, Q., Sun, L., Yang, F., Song, X., Gao, J., Wang, X., Xu, H.: Time series data augmentation for deep learning: A survey. In: Proceedings of the Thirtieth International Joint Conference on Artificial Intelligence. International Joint Conferences on Artificial Intelligence Organization (aug 2021). \doi{10.24963/ijcai.2021/631}, \url{https://doi.org/10.24963\%2Fijcai.2021\%2F631}

\bibitem{wen2023transformers}
Wen, Q., Zhou, T., Zhang, C., Chen, W., Ma, Z., Yan, J., Sun, L.: Transformers in time series: A survey (2023)

\bibitem{jiang2019resnet}
Wu, J., Zhang, Z., Ji, Y., Li, S., Lin, L.: A resnet with ga-based structure optimization for robust time series classification. In: 2019 IEEE International Conference on Smart Manufacturing, Industrial \& Logistics Engineering (SMILE). pp. 69--74 (2019). \doi{10.1109/SMILE45626.2019.8965287}

\bibitem{YANG2022108606}
Yang, X., Zhang, Z., Cui, R.: Timeclr: A self-supervised contrastive learning framework for univariate time series representation. Knowledge-Based Systems  \textbf{245},  108606 (2022). \doi{https://doi.org/10.1016/j.knosys.2022.108606}, \url{https://www.sciencedirect.com/science/article/pii/S0950705122002726}

\bibitem{zeiler2012adadelta}
Zeiler, M.D.: Adadelta: An adaptive learning rate method (2012)

\bibitem{zhumekenov2019fourier}
Zhumekenov, A., Uteuliyeva, M., Kabdolov, O., Takhanov, R., Assylbekov, Z., Castro, A.J.: Fourier neural networks: A comparative study (2019)

\bibitem{ziyin2020neural}
Ziyin, L., Hartwig, T., Ueda, M.: Neural networks fail to learn periodic functions and how to fix it. In: Larochelle, H., Ranzato, M., Hadsell, R., Balcan, M., Lin, H. (eds.) Advances in Neural Information Processing Systems. vol.~33, pp. 1583--1594. Curran Associates, Inc. (2020), \url{https://proceedings.neurips.cc/paper_files/paper/2020/file/1160453108d3e537255e9f7b931f4e90-Paper.pdf}

\end{thebibliography}

\appendix

\end{document}